\documentclass{segabs}
\usepackage{graphics}
\usepackage{amsmath}
\usepackage{mathtools}
\usepackage{amsfonts}
\usepackage{multirow}
\usepackage{booktabs}
\usepackage{amssymb}
\usepackage{array,ragged2e}
\newcolumntype{P}[1]{>{\RaggedRight\arraybackslash}p{#1}}
\usepackage{hyperref}
\usepackage[english]{babel}
\usepackage{blindtext}
\DeclareMathOperator*{\argmin}{arg\,min}

\begin{document}

\onecolumn

\date{}% Certain latex templates automatically add the compilation date as a footnote in the generated pdf. This command can remove the date in some of the templates but does not work for all.

\onecolumn % make sure you keep this coverpage as one column. In this template, we force the coverpage to be one column with this command and then switch to double column for the remaining of the paper with the \doublecolumn command. 

%\begin{description}[labelindent=-3cm,leftmargin=1cm,style=multiline]
\begin{itemize}
\item[\textbf{Citation}]{Mustafa, Ahmad, and Ghassan AlRegib. "Explainable machine learning for hydrocarbon prospect risking." Second International Meeting for Applied Geoscience \& Energy. Society of Exploration Geophysicists and American Association of Petroleum Geologists, 2022.}

\item[\textbf{DOI}]{\url{https://doi.org/10.1190/image2022-3752055.1}}

\item[\textbf{Review}]{Date of presentation: 31st August 2022}

\item[\textbf{Data and Code}]{Data is proprietary and can not be shared.}

\item[\textbf{Bib}] {@inproceedings\{mustafa2022explainable,\\
  title=\{Explainable machine learning for hydrocarbon prospect risking\},\\
  author=\{Mustafa, Ahmad and AlRegib, Ghassan\},\\
  booktitle=\{Second International Meeting for Applied Geoscience \& Energy\},\\
  pages=\{1825--1829\},\\
  year=\{2022\},\\
  organization=\{Society of Exploration Geophysicists and American Association of Petroleum Geologists\}
\}
}

% Preprint sharing policy can vary depending on the publisher. Before posting a paper to arXiv, please specifically check the transaction/convference you are targeting. In some transactions, papers are usually added to arxiv after acceptance. Pubslishers usually allow the authors to share accepted version of their papers but not the final formatted version that is provided by the pubisher. In case of sharing preprints, publishers usually ask to add DOI and citation to the paper along with a copyright notice.

\item[\textbf{Contact}]{\href{mailto:amustafa9@gatech.edu}{amustafa9@gatech.edu}  OR \href{mailto:alregib@gatech.edu}{alregib@gatech.edu}\\ \url{https://ghassanalregib.info/} \\ }
\end{itemize}

%Following command sequence was used to start the paper content from the following page and avoid numbering cover page.
\thispagestyle{empty}
\newpage
\clearpage
\setcounter{page}{1}

%Cover page was 1 column. \twocolumn changes the page format back to double column.
\twocolumn

\title{Explainable Machine Learning for Hydrocarbon Prospect Risking}
\renewcommand{\thefootnote}{\fnsymbol{footnote}} 

\author{Ahmad Mustafa\footnotemark[1]
  and Ghassan AlRegib, Center for Energy and Geo Processing (CeGP), School of Electrical and Computer Engineering, Georgia Institute of Technology, \{amustafa9,alregib\}@gatech.edu}

\footer{Example}
\lefthead{Mustafa \& AlRegib}
\righthead{Explainable Machine Learning for Hydrocarbon Prospect Risking}

\maketitle
\section{Summary}
Hydrocarbon prospect risking is a critical application in geophysics predicting well outcomes from a variety of data including geological, geophysical, and other information modalities. Traditional routines require interpreters to go through a long process to arrive at the probability of success of specific outcomes. AI has the capability to automate the process but its adoption has been limited thus far owing to a lack of transparency in the way complicated, black box models generate decisions. We demonstrate how LIME---a model-agnostic explanation technique---can be used to inject trust in model decisions by uncovering the model's reasoning process for individual predictions. It generates these explanations by fitting interpretable models in the local neighborhood of specific datapoints being queried. On a dataset of well outcomes and corresponding geophysical attribute data, we show how LIME can induce trust in model's decisions by revealing the decision-making process to be aligned to domain knowledge. Further, it has the potential to debug  mispredictions made due to anomalous patterns in the data or faulty training datasets. 

\section{Introduction}
Progress in machine learning research has led to major advancements for various computer vision and natural language processing tasks like image classification, object detection and localization, natural scene understanding, speech recognition, and machine translation, to name a few \citep{obj_detect,Imagenet,  scene_understanding, speech_recognition, singh2017machine}. It has also resulted in major breakthroughs in the extremely critical field of medical image analysis, helping in the detection of cellular structures, tissue segmentation, and in diagnosing various diseases \citep{deepmedical}. In exploration geophysics, it has now become a vital component in many seismic interpretation and inversion tasks, as evidenced by the plethora of works in in salt body delineation \citep{Wang2015, AsjadSaltDetection, AmirSaltDetection,haibinSaltbodyDetection}, fault detection \citep{haibinFaultDetection, HaibinFaultDetection2,Di2019Fault}, facies classification \citep{YazeedFaciesClassification, YazeedFaciesWeakClassification},  seismic attribute analysis \citep{Long2018TextureAnalysis, Di20183DcurvatureAnalysis, Alfarraj2018Multiresolutionanalysis}, structural similarity based seismic image retrieval and segmentation \citep{YazeedStructurelabelPrediction}, and seismic inversion \citep{Alfarraj2019_elastic, Mustafa2019, Mustafa2020Jointlearning}. 

Critical to a more widespread adoption of AI in geophysical routines and workflows is a deeper understanding of the model's reasoning process as it arrives at a particular decision. This is true of all areas heavy with medical, legal, ethical, and monetary concerns. It is now common knowledge that accuracy metrics alone seldom guarantee a model's reliability under unknown conditions. This can happen owing to several factors: a model could overfit to training data but perform poorly on test data; a distributional shift in the test data at run time could lead to significant degradation in model performance, especially if it has been trained on data of a different nature; or there could be spurious correlations present in the training data leading to the model learning to predict on false causal features. 

Hydrocarbon prospect risking \cite{TLE-DHI} is one such application requiring interpreters to gather data from multiple sources and modalities to process that and arrive at a decision: success or failure for a given oil prospect. A single wrong prediction could mean the loss of several million dollars. The DHI consortium aimed to gather best practices for the purposes of risking prospects by comparing and correlating the outcomes on drilled prospects from around the world to various features collected on those prospects. Given the extremely large number of feature data collected and interpreted for this purpose, the application stands to benefit greatly from adoption of AI---subject to addressing the concerns raised above.  

In this paper, we demonstrate how the adoption of a model-agnostic explanation technique leads us to better understand the reasoning process behind a variety of ML models as they arrive at decisions for prospect success or failure on a dataset of well outcomes and input features. The method called LIME (Locally Interpretable Model Explanations) works by fitting interpretable models in the neighborhood of datapoints of interest \citep{Lime-paper}. LIME does not assume any knowledge of the inner workings of the model being explained and works only by querying the model at and around individual datapoints to understand the local behavior of the decision boundary. As we demonstrate later, this helps to quantify trust in individual model predictions by correlating its reasoning process to prior geophysical knowledge and to debug models in case of faulty behavior. 

Our major contributions in this work are 1) training various machine learning models for the task of well outcome prediction for hydrocarbon risking, 2) using LIME to generate local explanations for various datapoints in the test set, and 3) identifying and verifying important correlations between features and well outcomes based on LIME explanations. To the best of our knowledge, this is the first work of its kind applying explainable machine learning to a geophysics application.   

\section{Theory}
Given a black box machine learning model $f$ and a datapoint $x$, LIME explains the prediction of $f$ over $x$ by fitting an interpretable model in the local neighborhood of $x$. The local neighborhood is denoted by $\pi_{x}$. The interpretable model, denoted by $g$, could be a logistic regression classifier or a decision tree. For an explanation to be plausible, the fitted model $g$ must be faithful to the decision boundary of the original model $f$ in the neighborhood of datapoint $x$. This property of the explanation is termed \emph{fidelity} and is measured by a loss term (denoted $L(f,g,x)$) penalizing the difference between the predictions over the datapoint $x$ for both $f$ and $g$. To better explore the decision boundary of $f$ in the region around $x$, LIME randomly samples new datapoints $x'$ in the vicinity of $x$ to obtain a new dataset ($Z$) containing both the original and sampled datapoints. It then queries the model $f$ to obtain their respective predictions. In the process to obtain the best fitted model $g$, the fidelity term incorporates predictions from all the datapoints contained in $Z$, the contribution from each being weighed according to its proximity to the original data point of interest $x$. The fidelity term is then more appropriately defined as $L(f,g,\pi_{x})$, where the latter refers to the fact that $g$ is being fitted to a dataset containing the original and sampled examples in the neighborhood $\pi_{x}$ and not just $x$ itself. Another property controlled by LIME for the generated explanations is the \emph{interpretability} of the fitted model $g$ in a term $\Omega(g)$. For the family of logistic regression classifiers, this could be a term penalizing the absolute norm of fitted coefficients (e.g., LASSO regression) to lead to more interpretable insights in the sparse weights. The complete optimization problem is then a sum of the objective functions measuring both fidelity and interpretability as shown in 

\begin{equation}
    \epsilon(x) = \argmin_{g\in G} \quad L(f,g,\pi_{x}) + \Omega(g),
    \label{eq:lime_global_formulation}
\end{equation}

where $\epsilon(x)$ refers to the explanation generated for the datapoint $x$ as the weights of the model $g$ (from a family of models $G$) that best minimizes the above cost function. A potential form $L$ could take is 

\begin{equation}
    L(f,g,\pi_{x}) = \sum_{z \in Z} \pi_{x}(z)(f(z) - g(z))^{2},
    \label{eq:form-L}
\end{equation}

where $z$ refers to the datapoints in set $Z$ (containing both the original and sampled data) and $\pi_{x}$ is a weighting function that weighs the contribution of each $z$ according to its distance from $x$. Points sampled closer to $x$ would contribute more to the loss function as opposed to those further away. An example of such a weighting function could be 

\begin{equation}
    \pi_{x}(z) = \text{exp}(-\frac{||x-z||^{2}}{\sigma}),   
\end{equation}

where $\sigma$ is a controlled hyperparameter. For more details, the reader is encouraged to refer to the work by \cite{Lime-paper}.

\section{Methodology}
The DHI consortium provides a dataset of drilled prospects around the world in a variety of geologic settings. For each prospect, it collects---among other things---data containing scores on how each prospect ranked on a variety of seismic amplitude anomalies, termed DHIs (Direct Hydrocarbon Indicators). The DHI risk assessment process as described in the work by \cite{TLE-DHI} computes the final Pg (Probability of Geologic success) in a three-step process incorporating 1) Initial Pg (estimated solely from geologic factors), 2) DHI Index (Pg estimated specifically from seismic amplitude anomalies associated with DHIs), and 3) data quality scores for various geophysical data. For all prospects pertaining to AVO class 3 geologic settings, we obtain their outcomes measured as either success (1) or failure (0). Also corresponding to each prospect, we obtain data on its initial Pg, DHI index, final Pg, and 3 other attributes measuring data quality for various geophysical modalities. These 6 attributes then form the feature data for each prospect in a six dimensional vector ($x\in \mathbb{R}^{6}$) whereas the prospect outcome forms the target class ($y\in[0,1]$). From the 258 such instances available, we randomly divide such prospects into training and test data with a 50/50 ratio. We then train a variety of machine learning models on the training split including a logistic regression classifier, support vector machines, and multilayer perceptrons. Afterwards, we use LIME as described in the section above to generate explanations for various datapoints in the test set. In the section below, we describe and elaborate on some of our findings.   

\section{Results}
\begin{figure}
    \centering
    \includegraphics[scale=0.5]{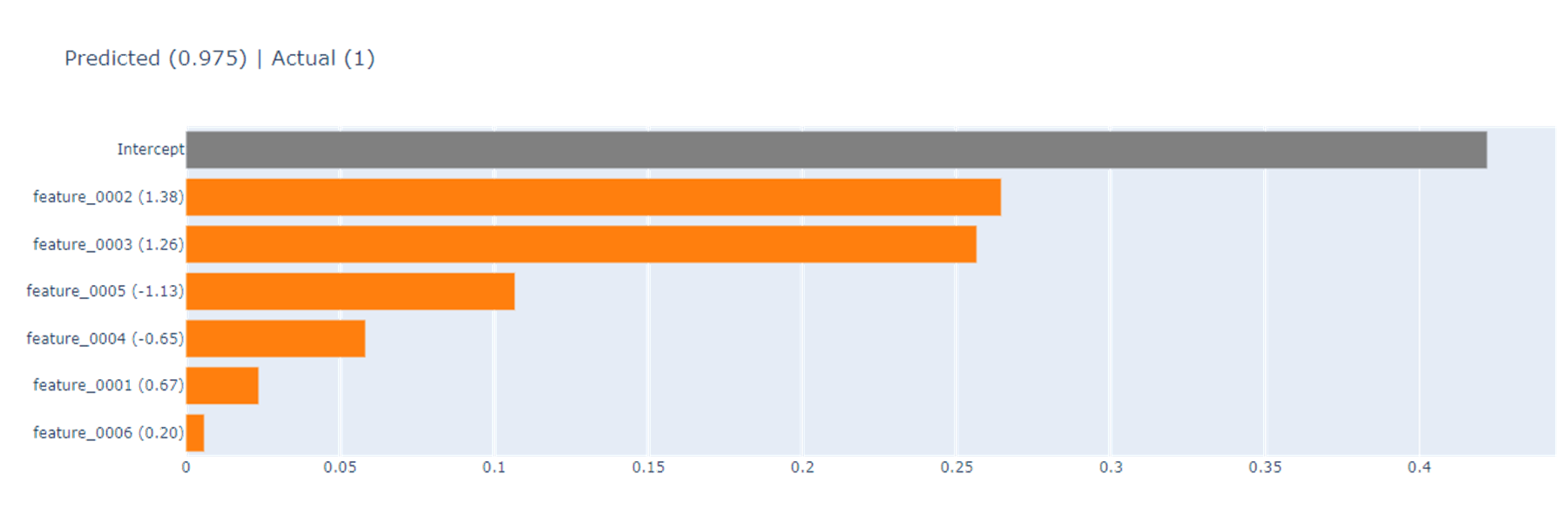}
    \caption{LIME explanations for logistic regression classifier. x-axis indicates weight values fitted to interpretable LIME models for various features. The values in parentheses next to feature names indicate their respective values.}
    \label{fig:lime_LR_strong_positive}
\end{figure}

\begin{figure*}
    \centering
    \includegraphics[scale=1]{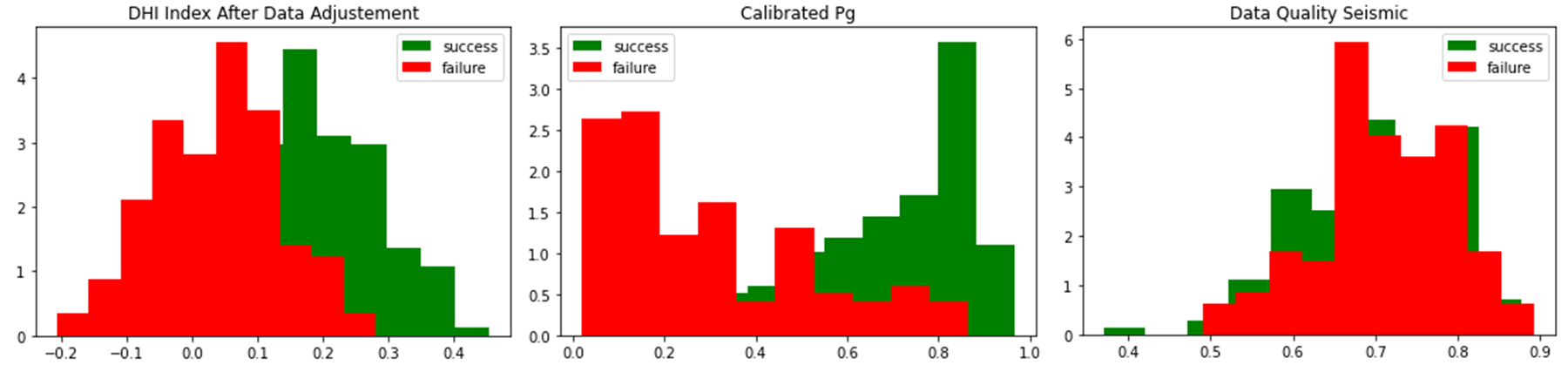}
    \caption{Plotting feature distributions for various geophysical markers in both success and failed prospects for DHI data.}
    \label{fig:DHI_feature_distributions}
\end{figure*}

Figure \ref{fig:lime_LR_strong_positive} depicts the LIME explanations generated for a certain datapoint in the test split for the logistic regression classifier. With the features as selected above, the classifier achieved training and test set accuracies of 0.84 and 0.77, respectively. This tells us the classifier was able to generalize to a reasonable extent. LIME is used to generate insights about the model's workings for specific data points. In this case, the classifier strongly predicts the prospect to be a success (with probability 0.975) where the ground truth is also success (class 1). In generating this prediction, it weighs positively all the 6 features present. This makes sense from a geophysical perspective since the better one scores on different criteria of the prospect risking process, the more likely one's chance of recovering hydrocarbons. Moreover, the explanations reveal that scoring high on features 2 and 3 (DHI index and Final Pg, respectively) is much more important in making this decision than for other features. This also makes sense from a geophysical point of view since these attributes generally play a very significant role in determining a particular prospect's chances of success. This finding is also corroborated by Figure \ref{fig:DHI_feature_distributions} where we plot the feature distributions for DHI Index, Final Pg, and Seismic data quality for both successes and failures in the DHI data. As can be observed, the former two features tend to be much more discriminatory compared to the last---something also verified by LIME. 

\begin{figure}
    \centering
    \includegraphics[width=\columnwidth]{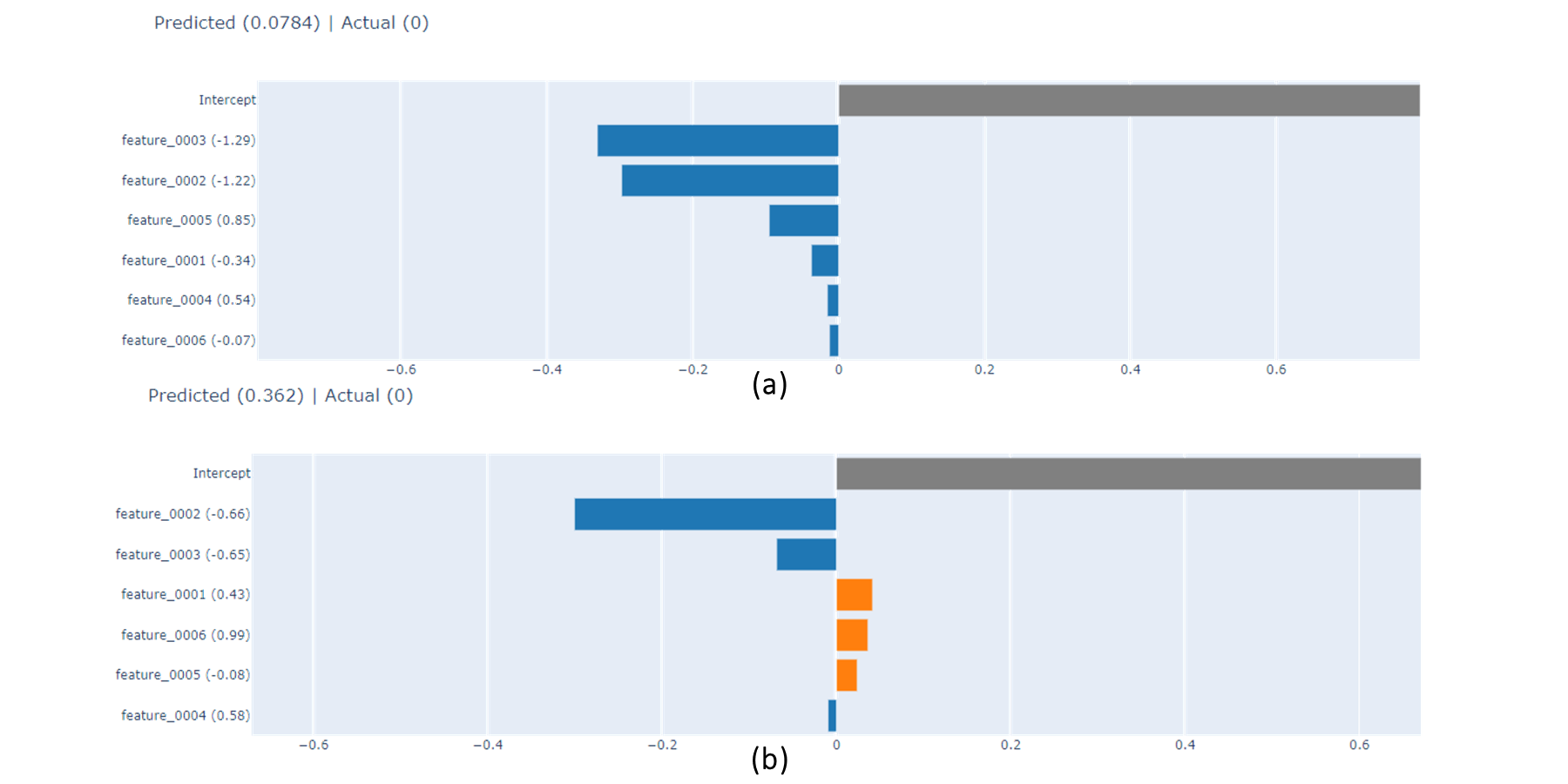}
    \caption{LIME explanations for failure class datapoints in test set for two cases: (a) strong negative and (b) a weak negative prediction.}
    \label{fig:lime_LR_strong_weak_negative}
\end{figure}

Figure \ref{fig:lime_LR_strong_weak_negative} further demonstrates the model's inner workings for two different test examples with negative ground-truths. In the first case (Figure \ref{fig:lime_LR_strong_weak_negative}a)), the model strongly predicts the prospect to be a failure, weighing negatively all features, but especially focusing on features 3 and 2 (Final Pg and DHI Index). Once again, this corresponds very well with what we know from prior geophysical knowledge. We also make a curious observation in Figure \ref{fig:lime_LR_strong_weak_negative}b); the models predicts a weak negative and the LIME explanations reveal it does so primarily on the back of weak values for features 2 and 3 but offsetting that with stronger values for features 1, 6, and 5 (Initial Pg and data quality scores), resulting in an overall weak negative.   

\begin{figure}
    \centering
    \includegraphics[width=\columnwidth]{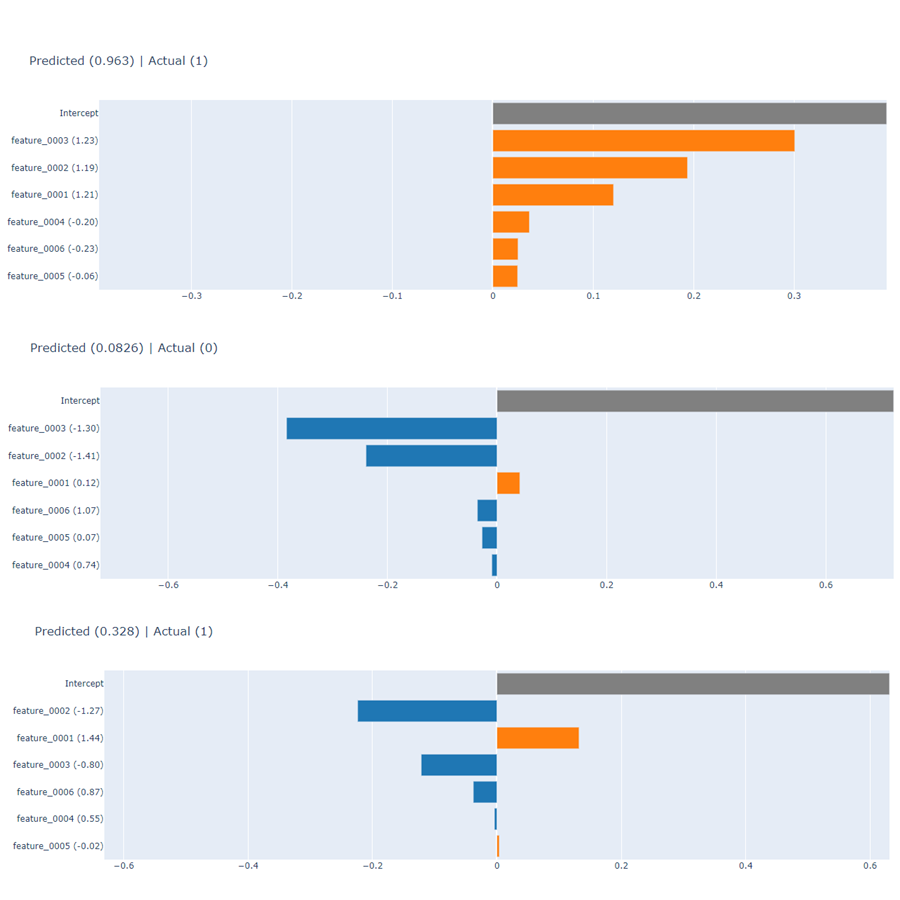}
    \caption{LIME explanations for SVM for a strong positive (top), a strong negative prediction (middle), and a weak negative wrongly predicted (bottom).}
    \label{fig:lime_svm_preds}
\end{figure}

We also generate LIME explanations similarly for an SVM model trained on a 50/50 split of the data between training and test sets. The results are depicted in Figure \ref{fig:lime_svm_preds}. Once again, we achieve reasonably high training and test accuracies of 0.84 and 0.75 respectively. Yet again, we see the SVM model fits positively to all features (but especially to final Pg and DHI index) to strongly predict positive a successful prospect. It underscores weak values in the same features to strongly predict a negative a prospect with negative ground truth. The bottom figure presents an interesting case study: the model predicts a weak negative against a successful prospect owing to relatively weak values for most features. It puts a positive weight on feature 1 (Initial Pg) despite common sense dictating otherwise (i.e., low initial Pg to predict a negative). For an interpreter, this would be a red flag for them to probe further into the particular prospect to see if there could be anomalous patterns present or to check for bugs in the model.    

\begin{figure}
    \centering
    \includegraphics[width=\columnwidth]{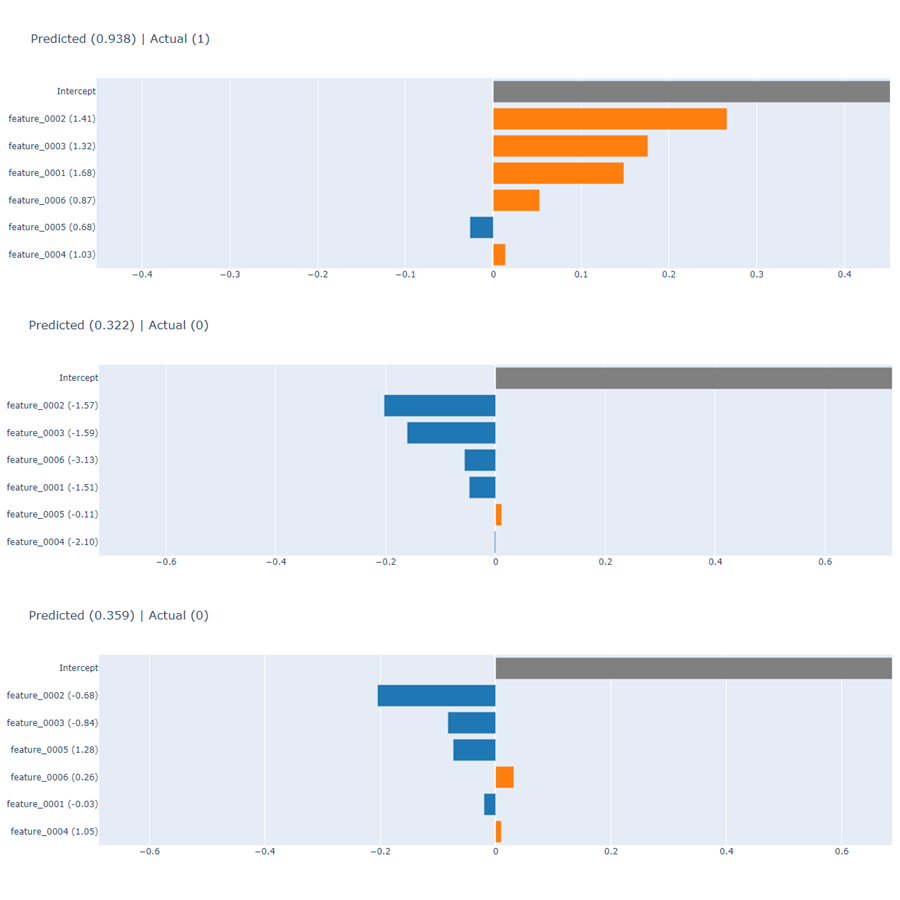}
    \caption{LIME explanations for an MLP for a strong positive (top) and two negatives (middle and bottom).}
    \label{fig:lime_mlp_preds}
\end{figure}

Figure \ref{fig:lime_mlp_preds} generates explanations for a two layer MLP acting on a positive and two negative examples from the test set. We obtain training and test accuracies of 0.83 and 0.81, respectively. As before, we observe the model to be learning to correlate primarily the DHI Index and Final Pg to the predicted result, although sometimes other features can come into play to weaken the prediction.    

\section{Conclusion}
Understanding the way complicated, black box models reason is central to widespread adoption of machine learning methods in areas requiring critical decision making having significant ethical, medical, legal, and monetary impacts. One way to assess trust in ML models is via model-agnostic techniques exploring their decision boundaries in local sub-regions around specific data points of interest. LIME generates explanations by fitting interpretable models to original and sampled data in these sub-regions. Applied to hydrocarbon prospect risking, LIME explanations can be used to verify model calibration by comparing its reasoning process to prior geophysical knowledge. Accuracy metrics may be misleading for a variety of reasons including but not limited to overfitting, data drift, data leakage, and spurious correlative features, among other reasons. Therefore, having qualitative interpretable explanations for model decisions over individual datapoints helps inject trust in model predictions, and hence, the model as a whole. 

By generating such explanations for different ML models for prospect risking, we verified the model reasoning process to broadly match an interpreter's by looking for the presence or absence of relevant features to generate predictions for prospect outcome. Our experiments revealed DHI Index and Final Pg to matter the most when deciding the outcome on a given prospect. We also observed how models at times made weak predictions owing to certain factors pulling the explanations in opposite directions (e.g., Figure \ref{fig:lime_LR_strong_weak_negative}b)). 

In addition to inducing trust in the model, the proposed workflow has the potential to be used to identify important causal factors and correlations in a much wider set of features than used for the experiments here. Moreover, such explanations can be used to perform feature selection to choose only the most relevant attributes for the task of prospect risking. By comparing explanations across multiple models, one can select for the most reliable model while still keeping testing/validation set accuracy in consideration.    

\bibliographystyle{seg}  % style file is seg.bst
\bibliography{refs}

\end{document}